\documentclass{article}
\usepackage{spconf,amsmath,graphicx}
\usepackage{amsfonts}
\usepackage{caption} 
\usepackage[table,xcdraw]{xcolor}
\usepackage[backend=bibtex,citestyle=numeric-comp,bibstyle=ieee,sorting=none,defernumbers=true,giveninits=true,doi=false,isbn=false,url=false,eprint=false,minbibnames=3,maxbibnames=4]{biblatex}
\usepackage{tabularx}
\usepackage{booktabs}
\usepackage{multirow}
\usepackage{ragged2e}
\usepackage{bm}

\usepackage{hyphenat}

\usepackage{boldline}

\newcolumntype{C}{>{\Centering\arraybackslash}X}

\newcolumntype{u}{>{\raggedright\hsize=.7\hsize}X}
\newcolumntype{t}{>{\Centering\hsize=.6\hsize}X}
\newcolumntype{s}{>{\Centering\hsize=.5\hsize}X}
\newcolumntype{o}{>{\Centering\hsize=.4\hsize}X}
\newcolumntype{k}{>{\Centering\hsize=.3\hsize}X}
\newcolumntype{y}{>{\Centering\hsize=.2\hsize}X}
\newcolumntype{z}{>{\Centering\hsize=.1\hsize}X}
\newcolumntype{v}{>{\raggedright\hsize=.6\hsize}X}
\newcolumntype{e}{>{\raggedright\hsize=.5\hsize}X}
\newcolumntype{j}{>{\raggedright\hsize=.35\hsize}X}
\newcolumntype{f}{>{\raggedright\hsize=.3\hsize}X}
\newcolumntype{h}{>{\raggedright\hsize=.2\hsize}X}
\newcolumntype{q}{>{\raggedright\hsize=.8\hsize}X}

\title{Learning Cross-lingual Visual Speech Representations}
\name{Andreas Zinonos, Alexandros Haliassos, Pingchuan Ma, Stavros Petridis, Maja Pantic}
\address{Department of Computing, Imperial College London, UK}
\begin{document}
%
\maketitle
\begin{abstract}
Cross-lingual self-supervised learning has been a growing research topic in the last few years. However, current works only explored the use of audio signals to create representations. In this work, we study cross-lingual self-supervised visual representation learning. We use the recently-proposed Raw Audio-Visual Speech Encoders (RAVEn) framework to pre-train an audio-visual model with unlabelled multilingual data, and then fine-tune the visual model on labelled transcriptions. Our experiments show that: (1) multi-lingual models with more data outperform monolingual ones, but, when keeping the amount of data fixed, monolingual models tend to reach better performance; (2) multi-lingual outperforms English-only pre-training; (3) using languages which are more similar yields better results; and (4) fine-tuning on unseen languages is competitive to using the target language in the pre-training set. We hope our study inspires future research on non-English-only speech representation learning.

\end{abstract}
\section{Introduction}
\label{sec:intro}

Cross-lingual learning uses data from different languages to create more robust and accurate models. Relevant supervised approaches \cite{burget2010multilingual, heigold2013multilingual, kannan2019large, ma2022visual} have delivered promising results, showing that increasingly larger models and training sets substantially improve performance. However, training such powerful models requires large amounts of labelled data, which can be time-consuming and expensive to acquire, hindering scalability.

Recently, several self-supervised learning approaches have been proposed which leverage large amounts of unlabelled data to learn better representations~\cite{he2020momentum, haliassos2022leveraging, grill2020bootstrap, devlin2018bert, baevski2019vq, baevski2019effectiveness, hsu2021hubert} and then fine-tune on the downstream tasks such as speech recognition, object detection, etc. Semi-supervised works which focus on various accents or dialects have also been released \cite{hu2021redat, li2018multi}. Some self-supervised approaches have also been proposed for learning visual speech representations, achieving promising results~\cite{DBLP:journals/corr/abs-2106-09171, DBLP:journals/corr/abs-2201-02184, haliassos2022jointly}. However, these works tend to use English-only data both for pre-training and fine-tuning, and it is not clear how they perform when non-English labelled or unlabelled data is used. A couple of self-supervised works~\cite{conneau2020unsupervised, babu2021xls} learn cross-lingual speech representations, but focus on improving the recognition of acoustic speech using multilingual models and do not compare monolingual models of the target language with the same pre-training dataset size. To the best of our knowledge, there is no existing work focusing on learning cross-lingual visual speech representations.

In this work, we aim to leverage a self-supervised method to learn visual representations for multiple languages. We use the RAVEn framework \cite{haliassos2022jointly}, where visual and audio encoders are pre-trained on unlabelled data via cross- and within-modal losses. The pre-trained visual backbone is then fine-tuned for the recognition of visual speech.

We make the following findings: (1) Pre-training on multiple languages leads to a lower character error rate (CER) compared to monolingual pre-training when more data is used, but to a higher CER when keeping the number of pre-training hours fixed; (2) multi-lingual pre-training outperforms English-only pre-training with the same number of data hours, showing that dataset size is not solely responsible for the improvement and that the information from other languages likely plays a significant role; (3) pre-training and fine-tuning on similar languages achieves better performance compared to less similar ones, signifying that the choice in languages is important; (4) fine-tuning on languages unseen during pre-training reaches competitive results compared to when including the language in the pre-training set.

\begin{figure}[!htb]
    \centering
    \centerline{\includegraphics[width=8.5cm]{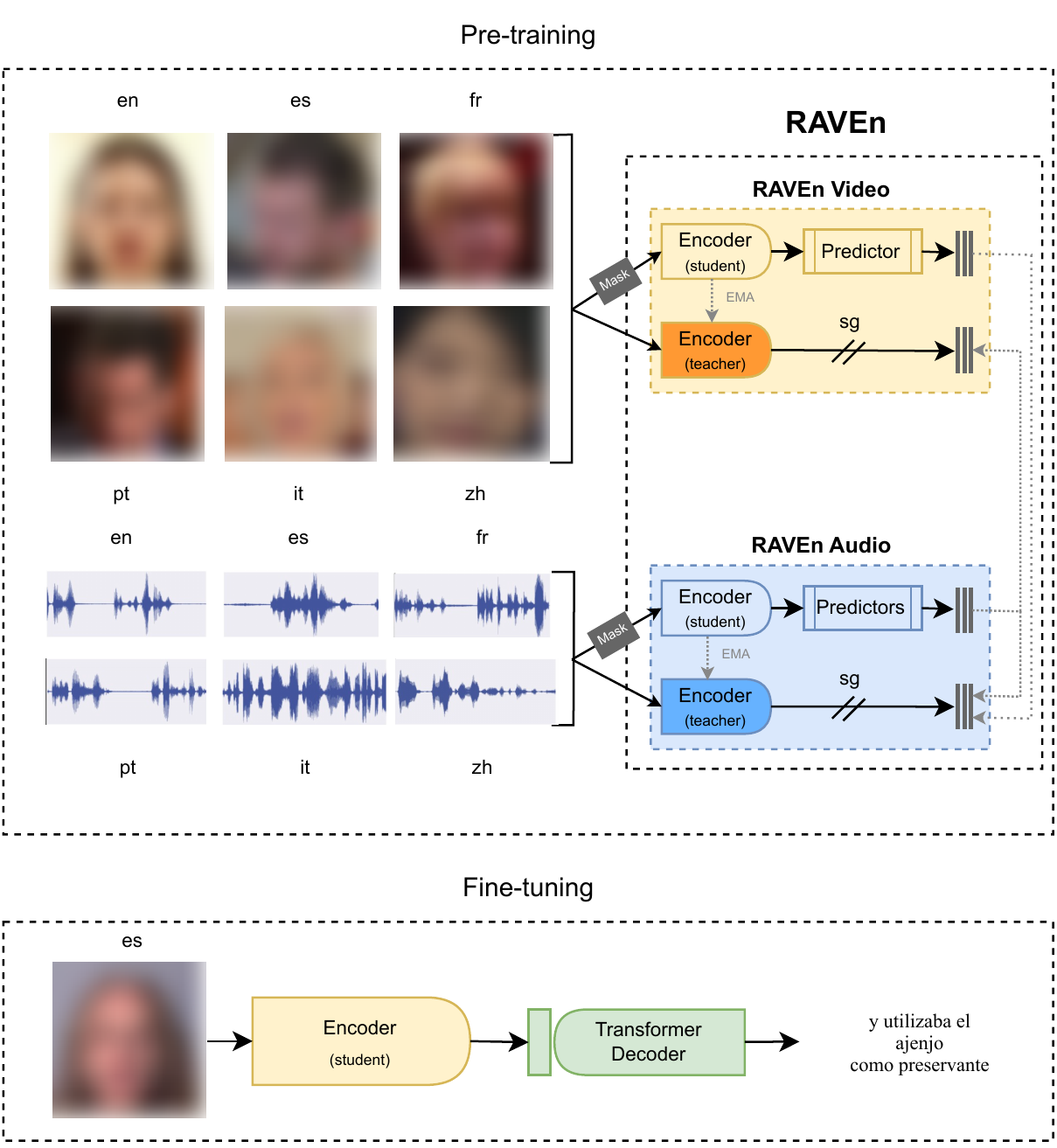}}
    \setlength{\belowcaptionskip}{-10pt}
    \caption{\textbf{Pre-training and fine-tuning pipelines}. We use RAVEn \cite{haliassos2022jointly} to pre-train models using audio and video from multiple languages. Stop-gradient and Exponential Moving Average are denoted by \textit{sg} and \textit{EMA}, respectively. We then initialise the visual-only encoder using the weights from the pre-training stage and fine-tune the entire network using the language-specific dataset. Frames blurred for anonymity.}
    \label{figure: RAVEn pipeline}
\end{figure}

\section{Datasets}
\label{sec:datasets}

We use four large-scale publicly-available audio-visual datasets for our study. For pre-training, we use the multilingual AVSpeech dataset \cite{ephrat2018looking}, while for fine-tuning we use LRS3~\cite{son2017lip} for English~(en), Multilingual TEDx~\cite{salesky2021multilingual} for Spanish~(es), Italian~(it), Portuguese~(pt) and French~(fr), and Chinese Mandarin Lip Reading (CMLR)~\cite{zhao2019cascade} for Mandarin~(zh).

LRS3 contains 438.9 hours of audio-video-text clips, with 408 hours of them in the pre-training set, 30 hours in the training-validation set and 0.9 hours in the test set. AVSpeech~\footnote{The statistics for the amount of hours for each language are calculated using the VoxLingua107 language classifier \cite{valk2021slt}.} comprises approximately 4\,700 hours of audio-visual segments from 290\,000 YouTube videos, with around 1\,333 hours in English, 337 hours in Portuguese, 204 hours in Spanish, 117 hours in French, 68 hours in Italian and 65 hours in Mandarin. Multilingual TEDx~\footnote{Videos where the targeted speaker is not visible are manually excluded.} contains 189 hours in Spanish, 189 hours in French, 164 hours in Portuguese and 107 hours in Italian, and the 86-hour CMLR dataset consists of 60.5 hours in the training set, 8.5 hours in the validation set and 17 hours in the test set.

\section{Method}
\label{sec:architectures}
\subsection{Pre-training}
\label{ssec:pre-training}
We use the audio-visual self-supervised RAVEn \cite{haliassos2022jointly} method for our study, as shown in Fig.~\ref{figure: RAVEn pipeline}. It consists of two pairs of student-teacher models, one for video and one for audio. The students intake masked inputs and predict targets generated by momentum-based teachers \cite{grill2020bootstrap}, which are given unmasked inputs. The video encoder consists of a ResNet-18 \cite{he2016deep} with a 3D stem~\cite{stafylakis2017combining}, followed by a Transformer encoder \cite{vaswani2017attention} with attention dimension 512, 8 heads, 2048 hidden units, and 12 layers. The audio encoder comprises a 1D ResNet-18~\cite{DBLP:conf/icassp/PetridisSMCTP18}, with the output piped into a Transformer \cite{gulati2020conformer} with the same settings as for the video encoder. The teachers have equivalent architectures to their student counterparts, apart from the students also containing additional 2-block Transformer predictors with attention dimension 512, 8 heads and 2048 linear units, which use mask tokens corresponding to the masked input. The predictors prevent representation collapse \cite{grill2020bootstrap}.

The audio student predicts both the video and audio targets, while the video student predicts the audio targets only. The loss between the predictions and targets is the cosine similarity. The students are optimised with a usual gradient-based method using the following loss: 
\begin{align}
\mathcal{L} = \text{sim}(p_s^a, \text{sg}(e_t^v)) + \text{sim}(p_s^v, \text{sg}(e_t^a)) + \text{sim}(p_s^a, \text{sg}(e_t^a)),
\end{align}
where sim is the cosine similarity function, sg the stop-gradient, $p_s^a$ and $p_s^v$ the student predictions, and $e_t^a$ and $e_t^v$ the teacher encodings for audio and video respectively. The teacher models are updated using an exponential moving average (EMA) of the student weights using the following equation: 
\begin{align}
\xi^m \leftarrow \tau\xi^m + (1 - \tau)\theta^m,
\end{align}
for momentum parameter $\tau$, modality $m\,\in\,\{v,\,a\}$, student weights $\theta$ and teacher weights $\xi$, with $\tau$ starting at 0.999 and following a cosine schedule to 1.

\subsection{Fine-tuning}
\label{ssec:fine-tuning}
Following \cite{haliassos2022jointly}, we fine-tune the pre-trained visual student encoder to perform visual speech recognition by attaching a linear layer and a Transformer decoder with attention dimension~256,~4 heads,~6 layers and 2048 linear units. We apply a joint CTC / attention loss~\cite{watanabe2017hybrid}. The beam size and CTC weight are fixed to 40 and 0.1 respectively, as in~\cite{ma2021end}. For labels, we use a character set for Mandarin and subword units~\cite{kudo2018sentencepiece} of vocabulary size 1000 for all other languages.

\section{Experiments}
\label{sec:experiments}

\subsection{Pre-processing}
\label{ssec:pre-processing}
Dataset pre-processing follows \cite{ma2022visual}. A 96$\times$96 bounding box is then used to crop a region around the speaker's mouth. Videos are then converted to grayscale and standardised by subtracting the mean and dividing with the standard deviation of the dataset. For audio the raw input is used without pre-processing or normalization. Utterances spanning longer than 24 seconds are split into smaller clips.

\subsection{Data augmentation}
\label{ssec:data-augmentations}
During training, we randomly crop the input video to a size of~88$\times$88 and perform horizontal flipping with probability 0.5~\cite{ma2022visual}. The transformations are applied throughout the temporal dimension of the clip consistently. No augmentations are performed for the audio.

\subsection{Masking}
\label{ssec:masking}
Our masking strategy in pre-training follows~\cite{haliassos2022jointly}: we perform random sampling on the video frames with probability~0.2 to select any as the start mask index, from where we zero out the next three frames. We apply a similar mask to the audio clip with a scaling factor of 640 to account for the sampling size difference compared to video.

\subsection{Training settings}
\label{ssec:training-settings}
We pre-train for 150 epochs with a peak learning rate of 3e-3, a linear warm-up for the first 40 epochs, and a cosine decay for the remaining epochs. The optimiser used is AdamW \cite{loshchilov2017decoupled} with $\beta_1 = 0.9$, $\beta_2 = 0.999$ and weight decay 0.04.  

We fine-tune for 50 epochs, with a 20-epoch warm-up and a cosine learning rate decay. We use a learning rate of 1e-3 for the encoder with layer-wise decay with parameter 0.5 \cite{clark2020electra}, and a learning rate of 5e-3 for the decoder. The optimiser is AdamW with $\beta_1 = 0.9$, $\beta_2 = 0.98$ and weight decay 0.1. 

\subsection{Fine-tuning}
\label{ssec:fine-tuning experiments}
We fine-tune on a single language at a time. A subset of LRS3 (``trainval'' partition) with a total of 30 hours is used to fine-tune in English. Multilingual TEDx is used to fine-tune in Spanish~(62\,h), French~(72\,h), Portuguese~(73\,h) and Italian~(40\,h). CMLR is used to fine-tune in Mandarin~(60.5\,h). We measure performance using Character Error Rate (CER) instead of Word Error Rate (WER), since Mandarin characters are not separated by spaces and to maintain consistency with the other languages.

\section{Results}
\label{sec:results}

\subsection{Multilingual pre-training surpasses monolingual}
\label{ssec:multilingual vs monolingual}

We pre-train (1) six monolingual models with either English, French, Italian, Spanish, Portuguese, or Mandarin~(``\textit{Monolingual}''); (2) one multilingual model with all six languages combined (``\textit{Multilingual}''); and (3) one with all languages combined but with the same number of hours as for the English-only dataset (``\textit{Multilingual RE}''), since AVSpeech contains more English-speaking videos than other languages\footnote{We combine all languages except English and then randomly add English samples until we reach 1\,333\,hours for the Reduced English (RE) model. This results in 542\,hours of English and 791\,hours of other languages.}. Table~\ref{table: complete language results} shows that for every language except English, multilingual pre-training significantly outperforms monolingual. We hypothesise that the extra data helps lower-resource languages by enabling the model to learn structures of speech shared across different languages. For English, which contains plenty of pre-training data, adding more languages may cause the representations to become less English-specific, in turn harming downstream performance.

\begin{table}[!htb]
\small
\centering
\renewcommand\arraystretch{1.0}
\caption{\textbf{Character Error Rate (CER) of visual-only models for different languages}. We pre-train models on either one language or six, and fine-tune on each language independently. English-only refers to the English monolingual model.}
\vspace{-1em}
\begin{tabularx}{.99\linewidth}{o  c c c c c c }
\toprule
\multirow{2}{*}{\textbf{\shortstack{Model}}} &\multicolumn{5}{c}{\textbf{Languages}}
    \\
& \textbf{en} & \textbf{it} & \textbf{pt} & \textbf{fr} & \textbf{es} & \textbf{zh} \\
\midrule
\textbf{\shortstack{Pre-Trained Data\\(Hours)}}        &1\,333        & 68          & 337         & 117         & 204         & 65 \\
\midrule
\textbf{\shortstack{Fine-Tuned Data\\(Hours)}} &30          & 40          & 73          & 72          & 62          & 60.5 \\
\midrule\midrule
English-Only              & \textbf{28.5} & 39.0 & 41.6 & 44.8 & 35.9 & 16.2  \\
\midrule
Monolingual                     & \textbf{28.5} & 76.9 & 43.5 & 71.9 & 47.2 & 22.6 \\
\midrule
Multilingual       & 31.5 & 38.0 & 40.0 & 43.6 & 35.3 & \textbf{15.8} \\
\midrule
Multilingual RE         & 32.8 & \textbf{35.1} & \textbf{38.7} & \textbf{43.3} & \textbf{32.8} & 15.9   \\
\midrule
\multirow{1}{*}{Multilingual MH} & 32.8 & 75.7 &48.3 &73.3 &49.6 & 24.4 \\
\bottomrule
\end{tabularx}
\setlength{\belowcaptionskip}{-5pt}
\vspace{-1em}
\label{table: complete language results}
\end{table}

\subsection{Multilingual pre-training surpasses English-only pre-training with the same number of hours}
\label{ssec:multilingual vs english-only}
Seeing that multilingual models perform better when pre-trained with more data (Section \ref{ssec:multilingual vs monolingual}), we investigate whether simply adding more data explains the improvement. We trained a monolingual \textit{``English-only''} model and a \textit{``Multilingual RE''} model both with 1333 hours, as explained in section~\ref{ssec:multilingual vs monolingual}. We fine-tune both models on all languages independently, and see that the \textit{``English-only''} model outperforms the other monolingual models, signifying that the number of training hours does matter. However the \textit{``Multilingual RE''} model outperforms the English-only model in every language except English, as seen in Table~\ref{table: complete language results}. This shows that the addition of more data in itself does not fully explain the performance boost: the information from various languages gives a further reduction in the error even when using the same dataset size.

\subsection{Monolingual pre-training surpasses multilingual with same hours}
\label{ssec: monolingual beats multilingual same hours}

We now compare multilingual and monolingual models, keeping the number of training hours constant, to disentangle the effects of extra data from the language variety (see Table~\ref{table: complete language results}). We train six multilingual models (``\textit{Multilingual MH}'') (Matched Hours), each with equal parts of every language to add up to the same number of hours used for the monolingual model. For every language except Italian, the monolingual model surpasses the multilingual one. We also consider a low-resource scenario, in which we randomly sample 30 hours of each language and train six new monolingual models, as well as a multilingual one with 30 hours of data in total, sampling 5 hours from each language. Table~\ref{table: low resource languages} shows that for every language in this context, monolingual outperforms multilingual pre-training.

This leads us to conclude that having sufficient hours of training data in the target language is important and directly substituting with equivalent data hours from other languages will worsen performance. Nonetheless, assuming that the amount of data from the target language is held constant, adding additional data from other languages improves representation learning, as discussed in Section \ref{ssec:multilingual vs monolingual}. 

\begin{table}[!t]
\small
\centering
\renewcommand\arraystretch{1.0}
\caption{\textbf{CER on low-resource languages}. We pre-train six monolingual models with 30 hours of data randomly sampled from each language, and one multilingual model with the same hours of pre-training data by sampling 5 hours from each language randomly. We compare the results in the case where the languages are low-resource.}
\vspace{-1em}
\begin{tabularx}{.99\linewidth}{o  c c c c c c }
\toprule
\multirow{2}{*}{\textbf{\shortstack{Model}}} &\multicolumn{5}{c}{\textbf{Languages}}
    \\
& \textbf{en} & \textbf{it} & \textbf{pt} & \textbf{fr} & \textbf{es} & \textbf{zh} \\
\midrule
\textbf{\shortstack{Pre-Trained Data\\(Hours)}}        & 30          & 30          & 30          & 30          & 30          & 30                \\
\midrule
\textbf{\shortstack{Fine-Tuned Data\\(Hours)}} & 30          & 40          & 73          & 72          & 62          & 60.5              \\
\midrule\midrule
\multirow{1}{*}{Monolingual} & \textbf{79.8}  & \textbf{79.1} & \textbf{75.1} & \textbf{75.7}  & \textbf{82.6} & \textbf{97.4}             \\
\midrule
\multirow{1}{*}{Multilingual} & 84.4 & 82.4 & 77.8 & 77.5 & 90.0 & 99.7             \\
\bottomrule
\end{tabularx}
\label{table: low resource languages}
\end{table}
\begin{table}[!t]
\small
\centering
\renewcommand\arraystretch{1.0}
\caption{\textbf{CER comparison when fine-tuning Spanish, pre-training on Spanish with added languages}. We pre-train models on three different pairs of languages without the use of any labelled data, and then fine-tune the visual model using a total of 62\,hours of labelled video clips in Spanish.}
\vspace{-1em}
\begin{tabularx}{.99\linewidth}{v e y}
\toprule
\textbf{Model} &\textbf{\shortstack{Pre-Trained Data\\(Hours)}} &\textbf{es}
\\
\midrule
Spanish              & 204                &47.2          \\\midrule
Spanish + Mandarin   & 204 + 65 = 269     &43.6  \\ \midrule
Spanish + French     & 204 + 65 = 269     &42.7  \\ \midrule
Spanish + Portuguese & 204 + 65 = 269     &\textbf{40.5} \\
\bottomrule
\end{tabularx}
\setlength{\belowcaptionskip}{-10pt}
\label{table: similar language results}
\end{table}

\subsection{Related language pre-training performs better}
\label{ssec:related vs unrelated languages}
We hypothesise that, keeping the number of hours fixed, pre-training on languages more related to the target language will lead to better results than pre-training on dissimilar languages. To test this hypothesis, we pre-train three different models: Spanish~(204\,h) combined with either Portuguese~(65\,h), French~(65\,h)\footnote{We randomly sample the Portuguese and French datasets until we reach 65\, hours.} or Mandarin~(65\,h), and fine-tune on Spanish, as shown in Table~\ref{table: similar language results}. By adding Mandarin for pre-training, we observe an absolute reduction of 3.6\,\% in CER. By replacing Mandarin data with French, a further improvement of 0.9\,\% can be observed. The CER is further reduced to~40.5\,\% with an absolute reduction of 2.2\,\% when using Portuguese instead. This is likely due to the fact that Portuguese is the most similar language to Spanish, which benefits the learning of visual speech representations for the language.

\begin{table}[!t]
\small
\centering
\renewcommand\arraystretch{1.0}
\caption{\textbf{CER on unseen languages}. We pre-train two multilingual models one excluding Portuguese and one Spanish, using the same number of pre-training hours as the Multilingual RE model we are comparing to.}
\vspace{-1em}
\begin{tabularx}{.99\linewidth}{v y y}
\toprule
\textbf{Model}  & \textbf{pt}  & \textbf{es} \\
\midrule
\textbf{Fine-Tuned Data (Hours)}             & 73          & 62   \\
\midrule\midrule
Multilingual RE & \textbf{38.7}   & \textbf{32.8}   \\
\midrule
Multilingual - Unseen (pt) & 41.5           & - \\
\midrule
Multilingual - Unseen (es) & -              & 34.0 \\
\bottomrule
\end{tabularx}
\vspace{-1em}
\label{table: unseen languages} 
\end{table}

\subsection{Representations on unseen languages}
\label{ssec: unseen languages}

To investigate how well the learned visual speech representations perform on unseen languages, we train two multilingual models with all languages except either Spanish or Portuguese, totalling 1\,333\,hours of pre-training data \footnote{We add English data to reach 1\,333\,hours when removing each language.}, the same as the ``\textit{Multilingual RE}'' model. We fine-tune on the unseen language, Portuguese and Spanish respectively. Results are shown in Table~\ref{table: unseen languages}. For comparison purposes, we use  ''\textit{Multilingual RE}'' model as our baseline. We show that the performance is competitive to the model for which the language was included in the data.

\section{Conclusion}
\label{sec:conclusion}
In this work, we investigated the impact of multilingual pre-training on visual speech recognition using a recently proposed self-supervised framework. We show that, on average, multilingual pre-training surpasses monolingual when pre-trained with more data by 23.8\% and 25.0\% in relative terms when comparing ``Monolingual'' with the ``Multilingual'' and ``Multilingual RE'' models respectively. However, it performs on average worse by 6.6\% CER compared to the ``Multilingual MH'', \textit{i.e.}, when fixing the dataset size. We also show that although using more pre-training data yields better results, training on a variety of languages results in a further improvement, yielding on average a relative CER decrease of 2.6\% (comparing the ``English-only'' to the ``Multilingual RE'' model). Furthermore, our experiments suggest that the similarity of languages when pre-training and fine-tuning is important, and that pre-training multilingual models yield competitive performance even when fine-tuning on unseen languages, with a relative performance decrease of 5.4\% on average (comparing the ``Multilingual RE'' model to ``Multilingual - Unseen'').

\clearpage
\section{References}
\begingroup
\setlength\bibitemsep{1pt}
\AtNextBibliography{\small}
\printbibliography[heading=none]
\endgroup

\end{document}